\pdfoutput=1
\documentclass[11pt]{article}

\usepackage[preprint]{acl}

 \usepackage{microtype}

\usepackage{times}
\usepackage[T1]{fontenc}
\usepackage[utf8]{inputenc}
\usepackage[english]{babel}
\usepackage[whole]{bxcjkjatype}

\usepackage{amsmath}
\usepackage{mathtools}
\usepackage{hyperref}
\usepackage{cleveref}
\usepackage{autonum}

\usepackage{booktabs}
\usepackage{multirow}

\usepackage{xspace}

\usepackage{leipzig}
\usepackage{gb4e}
\noautomath

\usepackage[inline]{enumitem}

\newcommand{\datasetname}{FoNLI\xspace}

\newleipzig{add}{add}{additive}
\newleipzig{posb}{posb}{possibility}
\newleipzig{npst}{npst}{non-past}
\newleipzig{exam}{exam}{exemplification}
\newleipzig{cls}{cls}{classifier}

\title{Analyzing Large Language Models on Japanese Focus Particle Inference}

\author{Yosuke Mikami \\
  The University of Tokyo \\
  RIKEN \\
  \texttt{ymikami@is.s.u-tokyo.ac.jp} \\\And
  Hitomi Yanaka \\
  The University of Tokyo \\
  RIKEN \\
  Tohoku University \\
  \texttt{hyanaka@is.s.u-tokyo.ac.jp} \\}

\begin{document}

\maketitle

\begin{abstract}
Japanese focus particles, such as \textit{mo} (``also'') and \textit{shika} (``only''), place semantic focus on a particular constituent against a set of alternatives, conveying meanings such as restriction, emphasis, or inclusion.
Consequently, sentences that are otherwise identical in their surface composition may receive diametrically opposed interpretations depending solely on the presence or absence of a focus particle.
Natural language inference (NLI) for such sentences is challenging, as models must grasp not only surface similarities but also subtle differences in meaning.
Furthermore, existing evaluation resources lack sufficient instances involving focus particles, making it difficult to evaluate how robustly models handle diverse lexical combinations and challenges like presupposition.
To address this, we introduce \datasetname, an NLI dataset specialized for Japanese focus particles, constructed by expanding an existing dataset through semi-automated generation and human validation.
Experiments on several LLMs show that the models 
1) achieve high accuracy on straightforward problems, 
2) demonstrate strong lexical robustness, and
3) exhibit well-calibrated prediction confidence.
However, all models struggle with presupposition and superficially similar but semantically distinct expressions, often making incorrect inferences with high confidence.
Analyses of thinking model outputs and internal representations suggest that these errors might be associated with an insufficient understanding of presupposition and an over-reliance on surface-level similarity.
\end{abstract}

\section{Introduction}
\label{sec:introduction}
Recent large language models (LLMs) have achieved remarkable performance across a wide range of natural language processing tasks~\citep{zhao2026survey}.
However, the evaluation of LLMs has primarily been conducted in English, leaving it unclear to what extent they can handle language-specific expressions and linguistic phenomena in other languages.
In addition, while it has long been noted that language models predict answers based on simple heuristics such as lexical overlap~\cite{mccoy-etal-2019-right,rajaee-etal-2022-looking}, it remains unclear whether similar issues persist in current LLMs.
To address these gaps, we take Japanese as a test case and analyze LLMs on natural language inference~(NLI;~\citealp{dagan2013recognizing,williams-etal-2018-broad}) involving Japanese focus particles.

Focus particles mark a particular constituent in a sentence as the focus, and contrast that constituent with a set of unstated alternatives.
In Japanese,  in the sentence \textit{Taro-mo ki-ta} (``Taro also came''), the focus particle \textit{mo} takes \textit{Taro} as its focus and the other relevant individuals in the context as its alternatives.
It contributes an additive meaning: in addition to some of the alternatives, Taro also came.

\begin{figure}
    \centering
    \includegraphics[width=\linewidth]{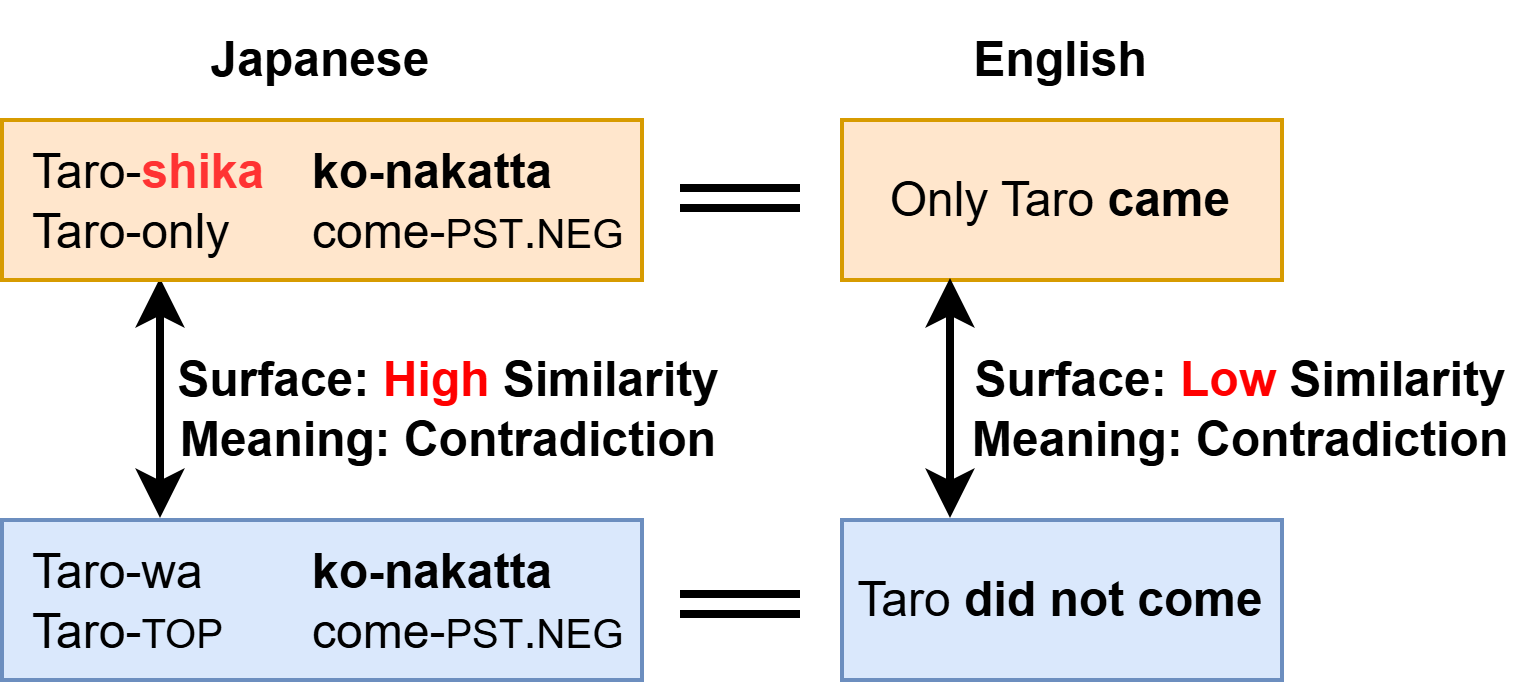}
    \caption{Example of challenging problems involving Japanese focus particles. Appendix~\ref{apd:gloss_abbreviations} explains the gloss abbreviations.}
    \label{fig:intro}
\end{figure}

Evaluating NLI involving Japanese focus particles poses two challenges.
First, focus particles can create sentence pairs that are superficially similar but semantically distinct.
As shown in the left part of \Cref{fig:intro}, \textit{Taro-shika ko-nakatta} (``Only Taro came'') and \textit{Taro-wa ko-nakatta} (``Taro did not come'') differ only in the particle attached to \textit{Taro}, forming a minimal sentence pair.
Crucially, the overt negation in the former does not negate the proposition but acts as a concord marker required by \textit{shika}, causing the two sentences to actually contradict each other.
In contrast, the English translation \textit{Only Taro came} lacks an overt negative morpheme, making it less superficially similar to \textit{Taro did not come}.
Judging NLI for such Japanese pairs thus requires models to look beyond surface-level lexical overlaps and accurately resolve fine-grained semantic nuances.

Second, existing evaluation resources lack sufficient coverage of focus particles to systematically test model robustness across diverse lexical items and complex semantic phenomena.
For instance, many focus particles trigger presupposition~\citep{potts2015presupposition}, which is an implicit assumption taken for granted.
Consider \textit{Taro-mo ki-ta kamoshirenai} (``Taro might have come too''), which presupposes that someone other than Taro came.
While operators like \textit{kamoshirenai} (modal of possibility) typically cancel standard entailments, the presupposition survives under such entailment-canceling operators (see \Cref{subsubsec:presupposition} for details).
Taken together, these challenges leave the question open of how robustly LLMs understand Japanese focus particles.

To address this gap, we construct an NLI dataset on Japanese focus particles, \datasetname\footnote{The dataset will be publicly available for research purposes at \url{https://github.com/ynklab/FoNLI}.}.
\datasetname is built from
a subset of the existing Japanese NLI dataset JSeM~\cite{10.1007/978-3-319-50953-2_5}.
We apply manual and LLM-based data augmentation and manually validate the judgments and sentence naturalness.
Compared with JSeM, FoNLI has greater lexical variation and a more balanced label distribution, allowing us to study prediction patterns and robustness of predictions for problems involving various combinations of vocabulary.

Using FoNLI, we evaluate multilingual and Japanese-specialized models.
We investigate not only the performance on NLI, but also whether models can robustly interpret the meaning of focus particles in inference problems involving diverse combinations of lexical items.
We further use token probability estimates as a proxy for model confidence to examine whether models make confident but incorrect predictions when solving challenging problems involving focus particles.

Our results show that the LLMs we studied
1) can generally solve straightforward task instances correctly, 
2) exhibit a high degree of robustness to lexical variation, and 
3) are generally less confident in incorrect predictions compared with in correct ones.
However, when it comes to the challenges described above, the models perform worse and frequently make highly confident errors.
Analyses of the outputs of LLM-jp-4-8B-Thinking~\citep{liu2026llm} and the internal representations of all models suggest that LLMs might not fully capture presuppositional meaning and rely excessively on surface-level similarity on our Japanese NLI task.

\section{Related Work}
\label{sec:related_work}
\subsection{Japanese NLI Dataset Involving Japanese Focus Particles}
\label{subsec:JSeM}
JSeM~\citep{10.1007/978-3-319-50953-2_5} is an existing Japanese NLI dataset that includes instances involving Japanese focus particles, manually constructed by experts in formal semantics.
There are 813 instances in the section on Japanese focus particles.
Each instance consists of three parts: one or more premises, a hypothesis, and a label, which takes one of the following values: \textit{yes}, \textit{no}, or \textit{unknown}\footnote{A small number of instances are annotated with labels other than these three (e.g., \textit{unacceptable}).}.
The label indicates whether the hypothesis can be inferred from the premises.

JSeM has several limitations for evaluating LLM performance.
First, JSeM typically contains only a single instance for each phenomenon or type of inference.
For example, the inference from \textit{Taro-mo repoto-o dasi-ta} (``Taro also submitted a report'') to \textit{someone other than Taro submitted a report} concerns the presupposition triggered by \textit{mo} when it is attached to the subject.
However, JSeM contains no other instances that test the same inference pattern.
As a result, JSeM is not well suited to evaluating LLMs in few-shot settings.
Furthermore, when a model fails on a given instance, JSeM does not allow us to determine whether the failure stems from an inability to understand the underlying inference pattern or from a lack of understanding of the specific phrases in that instance.

In addition, the label distribution in JSeM is substantially imbalanced: approximately 75\% of the instances in the focus particle section are labeled \textit{yes}.
This imbalance may hinder the analysis of model performance across different labels, particularly for instances with \textit{no} and \textit{unknown}.

Moreover, JSeM has not been validated by non-expert human annotators.
This could lead to the inclusion of instances that do not align with the intuitive judgments of non-expert native speakers.

These limitations motivate the need for a benchmark with greater lexical diversity, a more balanced label distribution, and non-expert validation.

\subsection{Analysis of Language Models through NLI}
\label{subsec:LLM_and_NLI}
Previous studies have used NLI to evaluate and analyze the reasoning capabilities of language models.
\citet{madaan-etal-2025-lost} argued that NLI has not yet reached benchmark saturation, since it still discriminates between stronger and weaker LLMs.
They compared the label distributions produced by LLMs and human informants using the Jensen-Shannon divergence~\citep{menendez1997jensen} and found a large divergence.
However, \citeauthor{madaan-etal-2025-lost} did not thoroughly investigate the relationship between model confidence and whether the predictions were correct.
\citet{wang-etal-2024-rethinking} reported that LLMs are often overconfident on NLI tasks, assigning high confidence to predictions that are incorrect.
However, they focused on standard, relatively simple NLI instances, and it remains unclear whether LLMs exhibit similar overconfidence on more challenging cases.
We therefore investigate this tendency in NLI involving Japanese focus particles using recent LLMs.

Several English NLI datasets~\citep{truong-etal-2022-another,jeretic-etal-2020-natural,francis-etal-2024-blame} include NLI problems involving focus-sensitive expressions such as \textit{only} and \textit{even}.
However, these datasets are not designed specifically to evaluate language models' understanding of focus particles.
For Japanese, several NLI datasets have been proposed to evaluate specific linguistic phenomena, such as temporal inference~\citep{sugimoto-etal-2023-jamp} and comparatives~\citep{mikami2025can}.
Existing datasets do not include instances involving Japanese focus particles.
We therefore cannot systematically evaluate LLMs' understanding of Japanese focus particles.
Moreover, evaluation in the existing literature has mainly focused on model performance; other aspects such as robustness and confidence have received little attention.
In contrast, we also evaluate models for robustness and confidence, providing a more comprehensive analysis of LLM behavior on Japanese NLI involving focus particles.

\section{Dataset Creation}
\label{sec:dataset_expansion}
We now describe the construction of \datasetname.
First, we start from a subset of JSeM.
We introduce new inference instances into the subset and balance the label distribution by adding instances with minor labels.
The instances obtained up to this step are defined as \textbf{base instances}.
We then increase the number of instances by substituting content words in the resulting base instances using an LLM that will not be evaluated in our experiments.
Finally, we manually validate the instances obtained in the previous step and filter out those deemed invalid.

To provide a base set of FoNLI, we first collect instances whose label is unique and is one of \textit{yes}, \textit{no}, or \textit{unknown} from the subset involving Japanese focus particles in JSeM.
This is because labels other than \textit{yes}, \textit{no}, and \textit{unknown} (e.g., \textit{unacceptable} or \textit{infelicitous}) fall outside the standard NLI formulation, and instances with multiple labels are ambiguous for evaluation.
This filtering leaves 508 instances for \textit{yes}, 72 instances for \textit{no}, and 89 instances for \textit{unknown}.

\subsection{Adding Instances for Previously Uncovered Properties}
\label{subsec:new_inference_type}
We manually design new instances that capture important properties of Japanese focus particles but are not covered by JSeM.
Due to space limitations, we present a subset of them below.
The other types of added instances are listed in Appendix~\ref{apd:added_problems}.

\subsubsection{Presupposition under Entailment-Canceling Operators}
\label{subsubsec:presupposition}
Presupposition~\citep{potts2015presupposition} is information whose truth is taken for granted.
Some of the focus particles covered by JSeM trigger presupposition.
One such particle is the additive particle \textit{mo} (``also'';~\citealp{shudo2013presupposition}).
For example, (\ref{ex:mo_presupposition_a}) presupposes (\ref{ex:mo_presupposition_b}).
\begin{exe}
    \ex\label{ex:mo_presupposition}
    \begin{xlist}
        \ex\label{ex:mo_presupposition_a}
            \gll Taro-\textbf{mo} ki-ta.\\
            Taro-also come-\Pst{}\\
            \glt ``Taro also came.''
        \ex\label{ex:mo_presupposition_b}
            \gll Taro igai-ni ki-ta hito-ga i-ru.\\
            Taro other.than-\Dat{} come-\Pst{} person-\Nom{} exist-\Npst{}\\
            \glt ``Someone other than Taro came.''
    \end{xlist}
\end{exe}
One important property of presupposition is that it is preserved under entailment-canceling operations, such as negation and modalization.
In the following examples (\ref{ex:mo_presupposition_cancel}), both sentences presuppose that someone other than Taro came, whereas the ordinary entailment that Taro came is canceled.
\begin{exe}
    \ex\label{ex:mo_presupposition_cancel}
    \begin{xlist}
        \ex\label{ex:mo_presupposition_cancel_a}
            \gll Taro-\textbf{mo} ki-ta \underline{toiuwakedewanai}.\\
            Taro-also come-\Pst{} \Neg{}\\
            \glt ``It is not the case that Taro also came.''
        \ex
            \gll Taro-\textbf{mo} ki-ta \underline{kamoshirenai}.\\
            Taro-also come-\Pst{} \Posb{}\\
            \glt ``Taro might have come too.''
    \end{xlist}
\end{exe}
JSeM does not contain instances targeting this property, and we cannot study whether LLMs understand the invariance of presuppositions under entailment-canceling operations.
We therefore add such problems, including one with (\ref{ex:mo_presupposition_cancel_a}) as the premise and (\ref{ex:mo_presupposition_b}) as the hypothesis.

Note that the extent to which a presupposition is shared among speakers may vary depending on the particle that triggers it.
For example, the presupposition of \textit{mo} is well-established in the literature on Japanese focus particles (e.g., \citealp{okabe-2001-semantics, shudo2013presupposition}).
In this respect, the presupposition of \textit{mo} can be regarded as a relatively robust presupposition.
In contrast, the same kind of presupposition by \textit{sae} is less clear.
While \citet{hasegawa2011semantics} analyzes \textit{Taro-sae ki-ta} (``Even Taro came'') as triggering the presupposition that someone other than Taro came, \citet{donati2025additivity} argue that \textit{sae} does not seem to be associated with the presupposition.
This disagreement suggests that the presupposition of \textit{sae} may be less robust than that of \textit{mo}, potentially giving rise to variation among speakers in their judgments.
We determine whether to include instances exhibiting inter-speaker variation in the dataset through human validation (see \Cref{sebsec:dataset_validation}).

\subsubsection{Interaction between Distributivity and Exclusivity}
\label{subsubsec:distributivity_exclusivity}
Distributivity is the property whereby, when an argument of a predicate denotes multiple individuals, the meaning of the predicate applies to each separately~\citep{champollion2019distributivity}.
For example, \textit{Taro and Jiro came} entails \textit{Taro came}.
This is because the predicate \textit{came} is interpreted distributively.

Distributivity sometimes interacts with exclusive focus particles, resulting in instances that cannot be solved by direct entailment alone.
This interaction can be exemplified by the following pair of premise~(\ref{ex:dist_excl_a}) and hypothesis~(\ref{ex:dist_excl_b}), both of which contain the exclusive focus particle \textit{dake} (``only'').
\begin{exe}
    \ex\label{ex:dist_excl}
    \begin{xlist}
        \ex\label{ex:dist_excl_a}
            \gll Taro to Jiro-\textbf{dake}-ga ki-ta.\\
            Taro and Jiro-only-\Nom{} come-\Pst{}\\
            \glt ``Only Taro and Jiro came.''
        \ex\label{ex:dist_excl_b}
            \gll Taro-\textbf{dake}-ga ki-ta.\\
            Taro-only-\Nom{} come-\Pst{}\\
            \glt ``Only Taro came.''
    \end{xlist}
\end{exe}
In this example, the hypothesis contradicts the premise.
In order to determine contradiction, a model needs to understand how \textit{dake} and distributivity function together in (\ref{ex:dist_excl_a}).
First, because the predicate is interpreted distributively across the conjunction \textit{Taro to Jiro}, the premise entails that Taro came and that Jiro came.
Second, the exclusivity of \textit{dake} in the hypothesis entails that no individual other than Taro came.
Consequently, the premise entails that Jiro came, whereas the hypothesis entails that no one other than Taro came, leading to a contradiction.

In this way, the process to determine the entailment relation between (\ref{ex:dist_excl_a}) and (\ref{ex:dist_excl_b}) consists of multiple steps, rather than relying on a single direct entailment.
We create an instance with (\ref{ex:dist_excl_a}) as the premise and (\ref{ex:dist_excl_b}) as the hypothesis.
We also generate three new instances from (\ref{ex:dist_excl_a}) and (\ref{ex:dist_excl_b}).
One new instance with the label \textit{no} is derived by replacing \textit{Taro} in (\ref{ex:dist_excl_b}) with \textit{Jiro}, and pairing the replaced sentence as a hypothesis with (\ref{ex:dist_excl_a}) as the premise.
The remaining instances are obtained by reversing the premise--hypothesis order of the original and the newly derived instances, respectively.

\subsection{Balancing the Dataset across Labels}
\label{subsec:balancing_label_dist}
After the procedures described in \Cref{subsec:new_inference_type}, the distribution of the gold labels is 559, 85, and 104 for \textit{yes}, \textit{no}, and \textit{unknown}, respectively.
The label distribution remains imbalanced.
We next describe how we increase the number of instances for \textit{unknown} and \textit{no}, respectively.

\begin{table*}[htbp]
    \centering
    \small
    \begin{tabular}{llll}
    \toprule
        ID & Premise & Hypothesis & Gold \\
    \midrule
        \begin{tabular}[t]{@{}l}
            \\ 219-1 \\
        \end{tabular} &
        \begin{tabular}[t]{@{}lll@{}}
            Taro-\textbf{mo} & repoto-o & dashi-ta.\\
            Taro-also & report-\Acc{} & submit-\Pst{}\\
            \multicolumn{3}{@{}l}{``Taro also submitted a report.''}
        \end{tabular}
        &
        \begin{tabular}[t]{@{}lll@{}}
            Taro-wa & repoto-o & dashi-ta.\\
            Taro-\Top{} & report-\Acc{} & submit-\Pst{}\\
            \multicolumn{3}{@{}l}{``Taro submitted a report.''}
        \end{tabular}
        &
        \begin{tabular}[t]{@{}l}
            \\  \textit{yes} \\
        \end{tabular}\\
    \midrule
        \begin{tabular}[t]{@{}l}
            \\ 550-3 \\
        \end{tabular} &
        \begin{tabular}[t]{@{}lll@{}}
            Amerika-\textbf{toka} & Kanada-\textbf{toka}-ni & iki-tai.\\
            America-such.as & Canada-such.as-\Dat{} & go-want\\
            \multicolumn{3}{@{}l}{``I want to go to places such as America and Canada.''}
        \end{tabular}
        &
        \begin{tabular}[t]{@{}lll@{}}
            Amerika-to & Kanada-ni & iki-tai.\\
            America-and & Canada-\Dat{} & go-want\\
            \multicolumn{3}{@{}l}{``I want to go to America and Canada.''}
        \end{tabular}
        &
        \begin{tabular}[t]{@{}l}
            \\  \textit{unknown} \\
        \end{tabular}\\
    \bottomrule
    \end{tabular}
    \caption{Examples in \datasetname. Focus particles are shown in bold.}
    \label{tab:dataset_examples}
\end{table*}

\subsubsection{Adding Variations of Monotonicity Inference}
\label{subsubsec:monotonicity}
In order to increase the number of \textit{unknown} instances, we utilize the monotonicity of logical entailment.
Monotonicity indicates how an entailment relation between two expressions is preserved or reversed. 
Replacing an expression with a more general one (e.g., a hypernym) preserves entailment, whereas the reverse substitution does not necessarily do so~\citep{icard2014recent,van1983determiners}.
JSeM already contains monotonicity-based instances, where a hypothesis is constructed by replacing a noun phrase in the premise with its hypernym or hyponym.
For example, the hypothesis in (\ref{ex:monotonicity_b}) is obtained from (\ref{ex:monotonicity_a}) by replacing \textit{freshman} with its hyponym \textit{freshman in the faculty of literature}.
\begin{exe}
    \ex\label{ex:monotonicity}
    \begin{xlist}
        \ex\label{ex:monotonicity_a}
            \gll \underline{ichinensei}-\textbf{sae} shussekishiteiru.\\
            freshman-even attend.\Prog{}\\
            \glt ``Even freshmen are attending.''
        \ex\label{ex:monotonicity_b}
            \gll \underline{bungakubu-no} \underline{ichinensei}-\textbf{sae} shussekishiteiru.\\
            faculty.of.literature-\Gen{} freshman-even attend.\Prog{}\\
            \glt ``Even freshmen in the faculty of literature are attending.''
    \end{xlist}
\end{exe}
However, the substitution patterns used in JSeM are limited.
Monotonicity relations can also be established through other types of substitutions, such as 
by adding prepositional phrases (e.g., \textit{person with a red hat} $\subseteq$ \textit{person}) or relative clauses (e.g., \textit{person who talked} $\subseteq$ \textit{person})~\citep{yanaka-etal-2019-neural}.

We use GPT-5.1\footnote{\url{https://openai.com/ja-JP/index/gpt-5-system-card}} to generate an example for each substitution pattern.
We then add only those examples judged to be natural to the dataset.

\subsubsection{Flipping the Polarity of Hypotheses}
\label{subsubsec:flipping_polarity}
Next, we increase the number of instances with \textit{no} labels.
In principle, for instances whose gold label is \textit{yes}, we can obtain instances with \textit{no} label by reversing the polarity of the hypothesis, that is, changing an affirmative hypothesis to a negative one, or vice versa.
We manually review each transformed sentence and exclude those judged to be semantically unnatural.

Hereafter, we refer to the instances obtained up to this step as \textbf{base instances}.
We have 1,041 base instances: 575 instances with \textit{yes},  321 with \textit{no},  and 145 with \textit{unknown}.

\subsection{Expanding the Variety of Content Words}
\label{subsubsec:variety_content_words}
To increase lexical diversity, we generate variants of existing premise--hypothesis pairs by rewriting content words while preserving their inference labels.
For each base instance obtained so far, we use GPT-5.1 to generate variants by changing only the content words.
Because the current dataset still has an imbalanced label distribution, we generate different numbers of variants for base instances with different labels.
Specifically, we generate 4 variants for each \textit{yes} instance, 7 for each \textit{no} instance, and 15 for each \textit{unknown} instance.
As a result, the dataset now has 2,875 instances for \textit{yes}, 2,568 for \textit{no}, and 2,320 for \textit{unknown}.

\subsection{Dataset Validation}
\label{sebsec:dataset_validation}
We conducted human validation of the dataset to filter out newly added instances that were unnatural or for which the annotators' judgments did not agree.

We constructed a validation dataset by randomly sampling two of its LLM-generated counterparts for each base instance.
Since there were 1,041 base instances, this procedure yielded a validation dataset of 2,082 instances.

The validation dataset was annotated by three crowdsourced workers, who were native speakers of Japanese and had no expertise in linguistics or semantics (see Appendix~\ref{apd:human_annotator} for details).
We asked the annotators to choose from four options: \textit{entailment}, \textit{contradiction}, \textit{neutral}, and \textit{invalid}.
The \textit{invalid} label was used for instances in which the sentences were ungrammatical or semantically unnatural.
The detailed definition of each label is shown in Appendix~\ref{apd:label_definition}.

For each base instance, if the majority vote assigned both sampled instances the same label as the gold label of the base instance, we retained all LLM-generated instances.
If the majority vote agreed with the gold label for only one of the two sampled instances, we manually inspected all instances derived from the corresponding base instance.
We discarded the other base instances and the LLM-generated variants.
This validation process yielded \datasetname, with 2,305 instances for the label \textit{yes}, 1,880 for \textit{no}, and 1,552 for \textit{unknown}.
Compared with the label distribution of our initial dataset~(508 for \textit{yes}, 72 for \textit{no}, and 89 for \textit{unknown}), \datasetname has a more balanced label distribution.
\datasetname also exhibits substantially greater lexical diversity.
JSeM contains 661 unique words, whereas \datasetname contains 2752.
Table~\ref{tab:dataset_examples} shows examples in \datasetname.

\section{Evaluation}
\label{sec:overall_evaluation}
We first evaluate the accuracy, robustness, and confidence of LLMs on \datasetname.
We conduct experiments in zero- and few-shot settings.
The zero-shot setting evaluates the overall accuracy and robustness of LLMs, whereas the few-shot setting examines how in-context examples affect them.
\label{subsubsec:zero-shot_setting}
\paragraph{Models}
We evaluate Llama-3.1-Swallow-[8B/70B]-Instruct~\citep{fujii2024continual} (hereafter Swallow-3.1-[8B/70B]), LLM-jp-4-8B-[Instruct/Thinking]~\citep{liu2026llm} (hereafter LLM-jp-4-8B-[It/Think]), Llama-3.1-[8B/70B]-Instruct~\citep{grattafiori2024llama} (hereafter Llama-3.1-[8B/70B]), and GPT-5.2\footnote{The URLs of the models are listed in Appendix~\ref{apd:model_url}.}.
Swallow-3.1-[8B/70B] are models obtained by continual pretraining of Llama-3.1-[8B/70B] on a Japanese corpus, and LLM-jp-4-8B-[It/Think] are trained from scratch on a large Japanese corpus.
We refer to Swallow-3.1-[8B/70B] and LLM-jp-4-8B-[It/Think] as Japanese-specialized LLMs.
We use Llama-3.1-[8B/70B] as baseline models and GPT-5.2 as a reference proprietary model.

\paragraph{Metrics}
We use accuracy to measure performance.
We propose Average Pairwise Agreement~(APA) as our robustness metric.
APA evaluates whether an LLM can robustly handle inference problems across different lexical expressions that preserve the underlying inference relation.
Specifically, APA is defined as the average of the following quantity computed for each base instance $b$.
\begin{equation}
    \frac{1}{
        \begin{pmatrix}
            \lvert P(b) \rvert \\
            2
        \end{pmatrix}
    }
    \sum_{\substack{p, p' \in P(b) \\ p \neq p'}}
    \mathbf{1}\left[ \operatorname{pred}(p) = \operatorname{pred}(p') \right]
\end{equation}
Here, $P(b)$ denotes the set of problem $b$ and its corresponding LLM-generated variants, and $\operatorname{pred}(p)$ denotes the label predicted by an LLM for problem $p$.
APA ranges from 0 to 1, with higher values indicating greater robustness to variation in content words.
Note that APA is independent of accuracy.

\paragraph{Prompt}
Although JSeM uses the labels \textit{yes}, \textit{no}, and \textit{unknown}, the standard NLI labels are \textit{entailment}, \textit{contradiction}, and \textit{neutral}.
To mitigate the impact of differences in label sets on performance, we use two types of prompts, each instructing the models to output one of these label sets.
We then take the average of the results obtained using the two prompts as the overall result.
The prompts are shown in Appendix~\ref{apd:prompt}.

\paragraph{Exemplars for Few-shot Experiments}
\label{subsubsec:oneshot_fewshot_settings}
We consider two settings: same-base and three-label.
In the same-base setting, for each test instance, we randomly sample an LLM-generated variant as the exemplar from the same base instance.
That is, the exemplar differs from the test instance only in its content words.
This setting aims to evaluate whether providing such an example helps the models correctly capture the inference pattern of focus particles.
In the three-label setting, for each test instance, we randomly sample one exemplar for each of the \textit{yes} (resp. \textit{entailment}), \textit{no} (resp. \textit{contradiction}), and \textit{unknown} (resp. \textit{neutral}) labels, each containing the same focus particle as the test instance.
The three exemplars are presented in random order.
This setting tests whether models can correctly distinguish the meaning of each label after providing few-shot examples of all three labels.

\begin{table*}[htbp]
    \centering
    \begin{tabular}{lcccccc}
    \toprule
    \multirow{2}{*}{Model} & \multicolumn{2}{c}{Zero-shot} & \multicolumn{2}{c}{Same-base} & \multicolumn{2}{c}{Three-label} \\
     & Accuracy & APA & Accuracy & APA & Accuracy & APA\\
    \midrule
    Llama-3.1-8B & .522 & .916              & .518 & .848               & .664 & .777 \\
    Llama-3.1-70B & .762 & .880             & .880 & .927               & .801 & .849 \\
    Swallow-3.1-8B & .721 & .864            & .931 & .912               & .584 & .561 \\
    Swallow-3.1-70B & .703 & .911           & \textbf{.976} & .965      & .815 & .845 \\
    LLM-jp-4-8B-It & .829 & \textbf{.961}  & .880 & \textbf{.962}      & .779 & \textbf{.898} \\
    LLM-jp-4-8B-Think & .794 & .937         & .708 & .794               & .617 & .627 \\
    GPT-5.2 & \textbf{.851} & .927          & .899 & .921               & \textbf{.849} & .871 \\
    \bottomrule
    \end{tabular}
    \caption{Accuracy and APA of each model in the zero-shot, same-base, and three-label settings. Each value indicates the average of the results obtained when each model was prompted to provide either JSeM-compliant labels (\textit{yes}, \textit{no}, \textit{unknown}) and when it was prompted to provide standard NLI labels (\textit{entailment}, \textit{contradiction}, \textit{neutral}). Boldface indicates the numerically highest value in each column.}
    \label{tab:zero-shot_results}
\end{table*}

\subsection{Results of Zero-shot Experiments}
\label{subsec:zero-shot}
\label{subsubsec:zero-shot_results}
We present the results of zero-shot experiments in the \textit{Zero-shot} column of \Cref{tab:zero-shot_results}.
All models achieve high APA scores of at least 0.864, suggesting that their predictions were highly robust to changes in content words.
In terms of accuracy, all models correctly predicted the gold labels, particularly for straightforward instances such as 219-1 in \Cref{tab:dataset_examples}.
GPT-5.2 reached the numerically highest performance~(0.851, compared to the other scores below 0.829).
Among the open-source models, LLM-jp-4-8B-It achieved APA and accuracy comparable to those of GPT-5.2.
Swallow-3.1-8B also showed relatively high APA~(0.911) and accuracy~(0.703).
Although Llama-3.1-8B exhibited a notably high APA~(0.916) relative to its accuracy~(0.522), this was primarily because it predicted \textit{yes} for the majority of the problems.

We further investigate LLMs' performance on each individual instance.
All models other than Llama-3.1-8B showed low accuracy on the 1,057 instances requiring presupposition reasoning.
\Cref{tab:presupposition_zeroshot} shows the accuracy and APA of each model on instances involving presupposition.
In particular, Swallow-3.1-70B, LLM-jp-4-8B models, and GPT-5.2 exhibited accuracy drops of more than 50\%.
Notably, accuracy was also low on instances involving the presupposition of \textit{mo}, which is relatively robust.
A breakdown of the results by focus particles is provided in Appendix~\ref{apd:presupposition_per_particle}.
To better understand these errors, we examined the reasoning outputs of LLM-jp-4-8B-Think (see Appendix~\ref{apd:thinking_outputs} for details).
We observed that the model frequently failed to infer presupposed content and occasionally treated presuppositions as being canceled by modal expressions.
These findings suggest that the observed performance degradation is largely attributable to an insufficient understanding of presupposition.
\begin{table}[tbp]
    \centering
    \begin{tabular}{lcc}
    \toprule
    Model & Accuracy & APA\\
    \midrule
    Llama-3.1-8B & .536 & .864 \\
    Llama-3.1-70B & .376 & .768 \\
    Swallow-3.1-8B & \textbf{.595} & .676 \\
    Swallow-3.1-70B & .257 & .826 \\
    LLM-jp-4-8B-It & .340 & \textbf{.915} \\
    LLM-jp-4-8B-Think & .306 & .884 \\
    GPT-5.2 & .415 & .783 \\
    \bottomrule
    \end{tabular}
    \caption{Accuracy and APA of each model on problems involving presupposition in the zero-shot setting. Each value represents the average of the results obtained when each model was prompted to provide either JSeM-compliant labels (\textit{yes}, \textit{no}, \textit{unknown}) and when it was prompted to provide standard NLI labels (\textit{entailment}, \textit{contradiction}, \textit{neutral}). }
    \label{tab:presupposition_zeroshot}
\end{table}

Poor performance in presupposition may also negatively affect performance in downstream tasks such as question-answering (QA) tasks (see Appendix~\ref{apd:downstream}).

A further manual inspection reveals that performance tended to be worse on instances where the premise and the hypothesis are superficially similar but the premise does not entail the hypothesis.
For example, in instances involving \textit{shika}, which is described in \Cref{sec:introduction}, larger models and LLM-jp-4-8b generally predicted the correct labels, whereas smaller models often failed to do so.
In addition, all models struggled with sentence pairs that are semantically similar but do not have an entailment relation, such as 550-3 in \Cref{tab:dataset_examples}.

\begin{figure*}
    \centering
    \begin{minipage}[t]{0.50\columnwidth}
        \centering
        \includegraphics[width=\columnwidth]{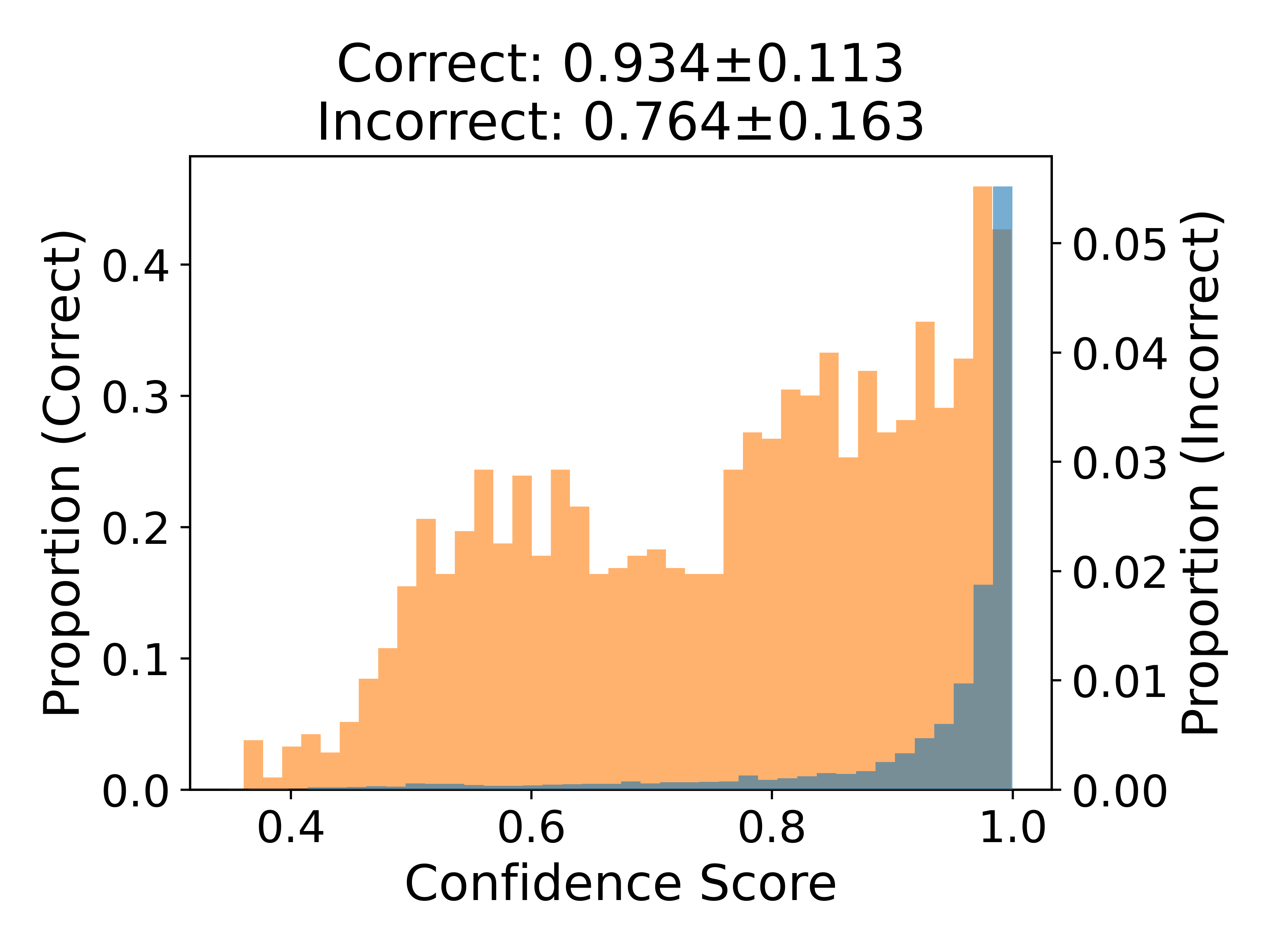}
        \caption*{Swallow-3.1-8B}
    \end{minipage}
    \begin{minipage}[t]{0.50\columnwidth}
        \centering
        \includegraphics[width=\columnwidth]{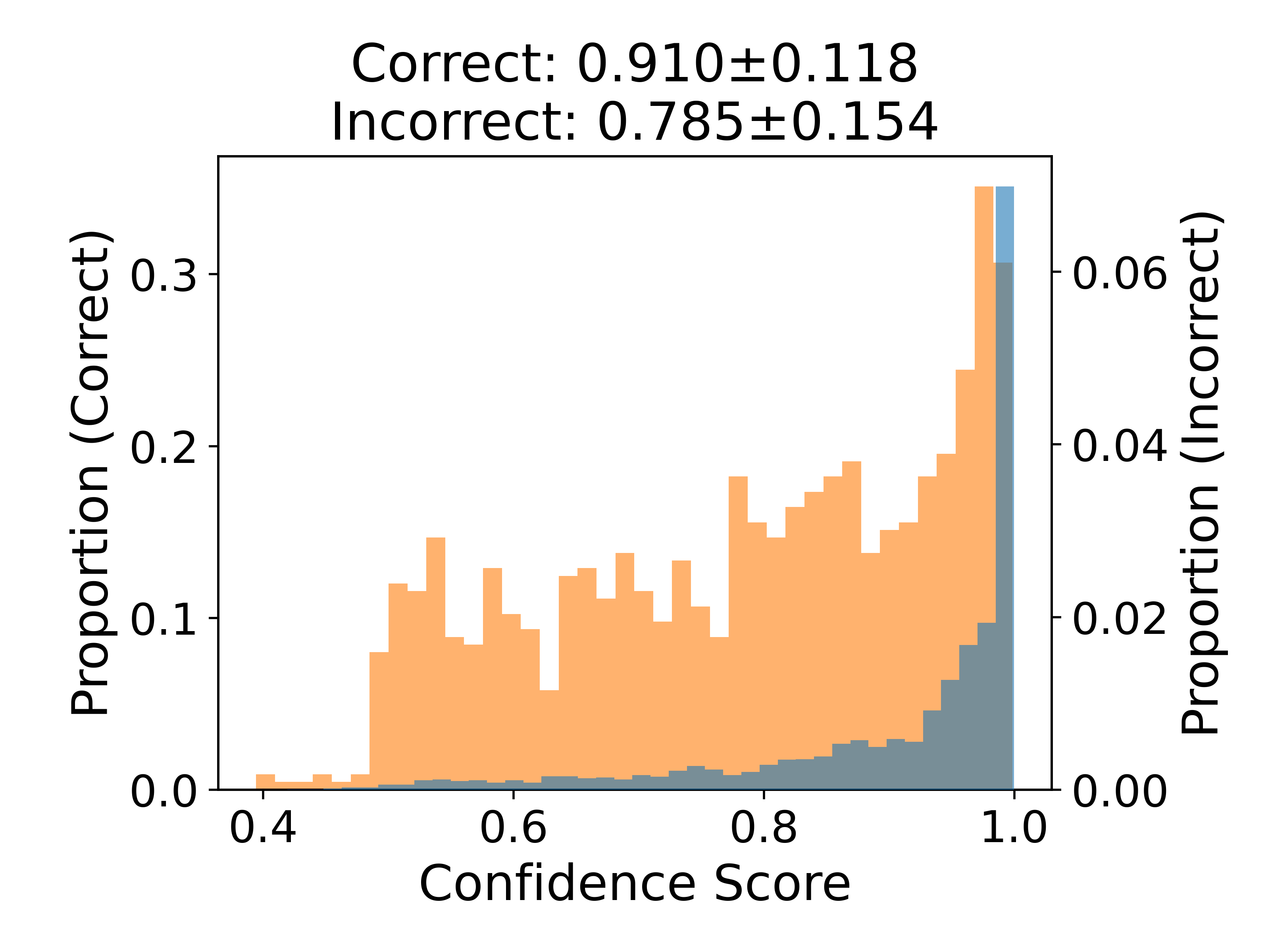}
        \caption*{LLM-jp-4-8B-It}
    \end{minipage}
    \begin{minipage}[t]{0.50\columnwidth}
        \centering
        \includegraphics[width=\columnwidth]{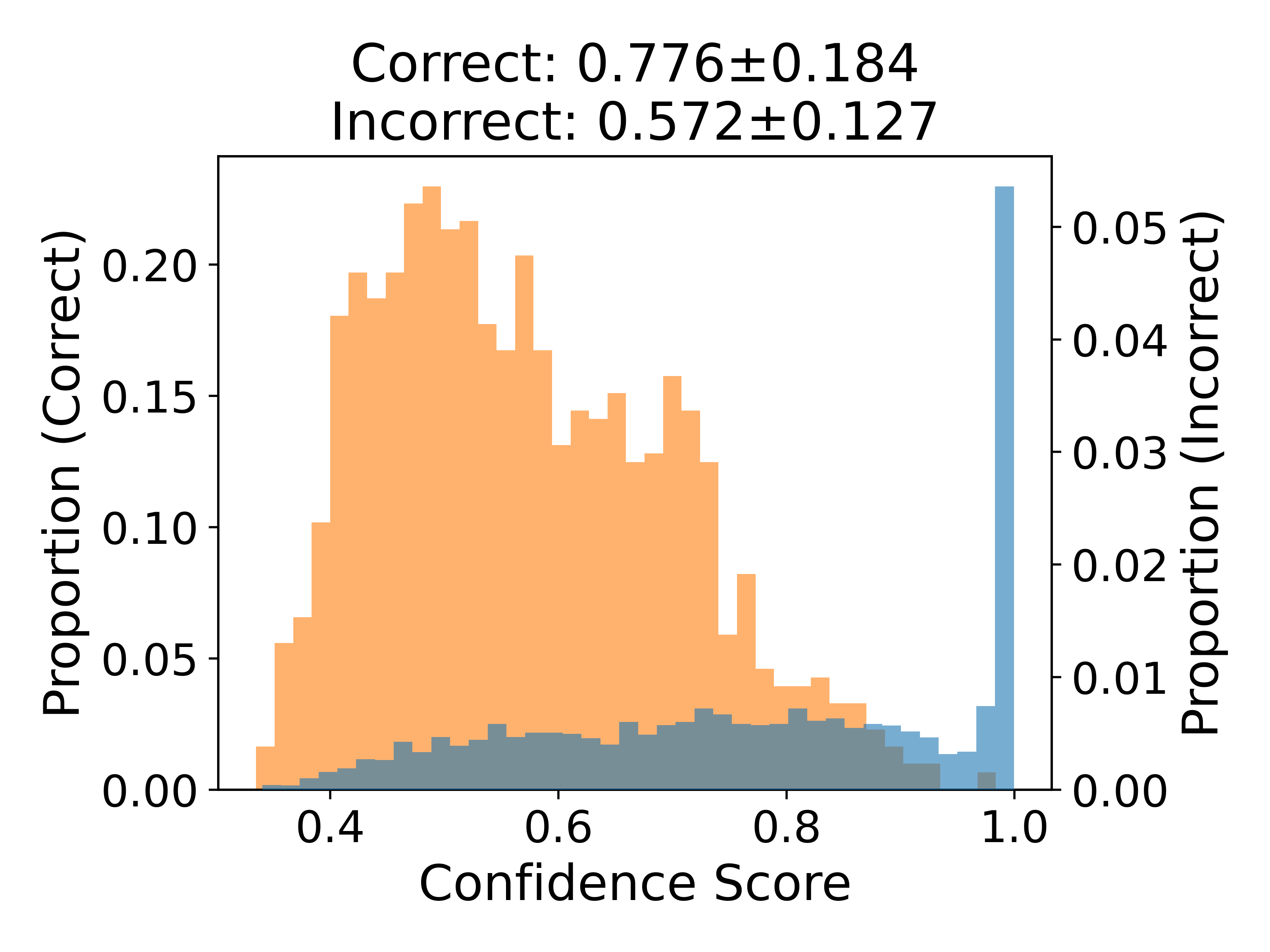}
        \caption*{Llama-3.1-70B}
    \end{minipage}
    \begin{minipage}[t]{0.50\columnwidth}
        \centering
        \includegraphics[width=\columnwidth]{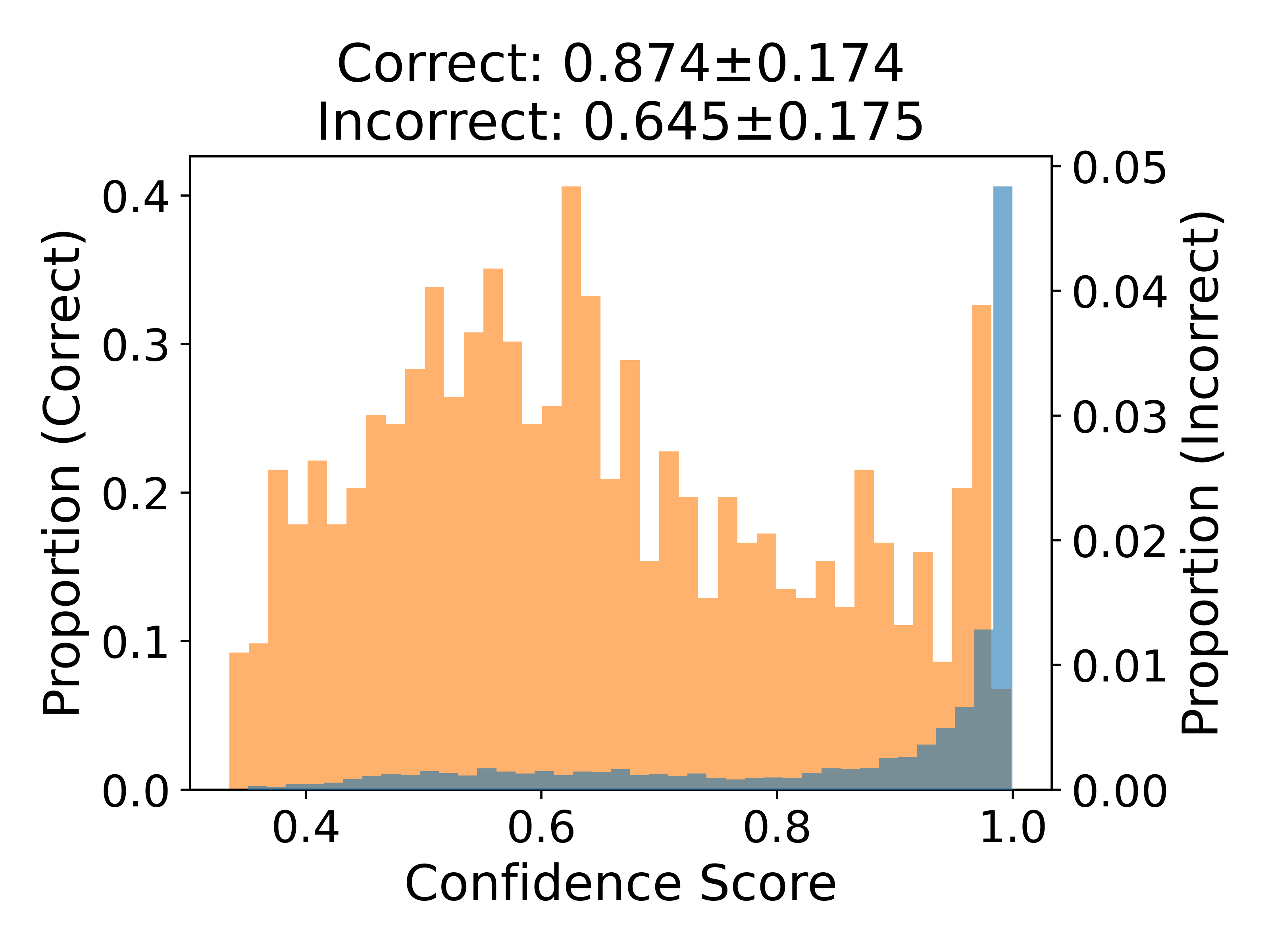}
        \caption*{Swallow-3.1-70B}
    \end{minipage}
    \caption{Distribution of prediction confidence for four models. The blue and orange distributions correspond to correctly predicted and incorrectly predicted instances, respectively. The values shown in the title of each plot indicate the mean and standard deviation of the prediction confidence.}
    \label{fig:logit_confidence}
\end{figure*}

\subsection{Results of Few-shot Experiments}
\label{subsec:oneshot_fewshot}

\label{subsubsec:oneshot_fewshot_results}
The results are shown in the \textit{Same-base} and \textit{Three-label} columns of \Cref{tab:zero-shot_results}.
Under the same-base setting, all models except for LLM-jp-4-8B-Think showed improved accuracy, with the improvements being particularly pronounced for the Swallow-3.1 models (over 0.9).
APA also increased for all models except Llama-3.1-8B and GPT-5.2.
In contrast, under the three-label setting, all models other than Llama-3.1-8B either showed lower accuracy than in the zero-shot setting or exhibited only marginal improvements over the same-base setting.
A similar trend was observed for APA, with all models except Llama-3.1-70B exhibiting a substantial decline.
One possible explanation is that providing few-shot examples of all three labels is insufficient for models to reliably distinguish their meanings and may even undermine the robustness of their predictions.

These results indicate that, in both prediction accuracy and APA, even models achieving high performance in the zero-shot setting remain sensitive to exemplars.

\subsection{Logits as Confidence Scores}
\label{subsec:logit_confidence}
We now study whether models make incorrect predictions with high confidence.
We take Llama-3.1-70B, Swallow-3.1 models, and LLM-jp-4-8B-It and use the logits as a proxy measure to assess their prediction confidence.
We do not include Llama-3.1-8B because its accuracy in the zero-shot setting is relatively low.
\label{subsubsec:logit_confidence_settings}
We extract the pre-softmax logits of \textit{yes}, \textit{no}, and \textit{unknown} at the final decoding position.
We then apply the softmax function to these logits, resulting in a probability distribution over the three labels.
We use the maximum of the resulting probabilities as \textbf{confidence scores}.

\subsubsection{Results and Discussions}
\label{subsubsec:logit_confidence_results}
\Cref{fig:logit_confidence} shows the distributions of confidence scores for correct and incorrect predictions.
Across models, the confidence score distribution for correct predictions is skewed toward higher values.
In contrast, the confidence score distribution for incorrect predictions has greater mass at lower values.
Particularly, larger models~(Llama-3.1-70B and Swallow-3.1-70B) made relatively few incorrect predictions with high confidence.

By inspecting the incorrect predictions with high confidence scores, we found instances involving presupposition and superficially similar sentence pairs in which the premise does not entail the hypothesis, both of which were challenging for the models in the zero-shot setting.
Among the incorrect predictions made with confidence scores above 0.95 (262, 96, and 214 instances for Swallow-3.1-[8/70]B and LLM-jp-4-8B-It, respectively), more than 60\% were accounted for by these properties.
This proportion exceeded 90\% for Swallow-3.1-70B and LLM-jp-4-8B-It.

These results show that the models are generally confident when making correct predictions and tend to be less confident when making incorrect ones.
However, this tendency does not hold for particularly challenging phenomena for which they often make incorrect predictions with high confidence.
Furthermore, a supplementary analysis (see Appendix~\ref{apd:conf_and_edl} for details) shows that higher-confidence predictions generally become stable in earlier layers, suggesting that model confidence is associated not only with prediction correctness but also with earlier commitment during inference.

\section{Possible Reliance on Surface Similarity}
\label{subsec:surface_similarity}
As discussed in \Cref{subsec:zero-shot}, several models in the zero-shot setting exhibited prediction patterns suggesting that they may confuse semantically distinct but surface-similar sentences.
In this section, we examine sentence embeddings produced by these models and investigate whether this tendency is reflected in their internal representations.

\subsection{Settings}
\label{subsubsec:surface_similarity_settings}
\paragraph{Similarity Measurement}
For Llama-3.1, Swallow-3.1, and LLM-jp-4-8B-It, we obtain sentence embeddings using PromptEOL~\citep{jiang-etal-2024-scaling}.
We insert each sentence into the following template, translated into Japanese from the template proposed by~\citet{jiang-etal-2024-scaling}, instructing the model to summarize its meaning in a single word.
\begin{center}
    この文「$\langle \text{sentence} \rangle$」の意味を一語で表すと：「\\
    (This sentence ``$\langle \text{sentence} \rangle$'' means in one word: ``)
\end{center}
We then input the resulting prompt to the model and use the hidden state of the next token as the sentence embedding.
We calculate the cosine similarity between the sentence embeddings.

\paragraph{Target}
\newcommand{\simsn}{\operatorname{sim}_{\mathit{SN}}}
\newcommand{\simsd}{\operatorname{sim}_{\mathit{SD}}}
As described in \Cref{sec:introduction}, sentences containing \textit{shika} (e.g., \textit{Taro-shika ko-nakatta} (``Only Taro came''); hereafter, \textit{shika} sentence) are superficially similar to their corresponding negative sentences without \textit{shika} (e.g., \textit{Taro-wa ko-nakatta} (``Taro did not come'')), although they contradict each other.
In contrast, sentences containing \textit{dake} (e.g., \textit{Taro-dake ki-ta} (``Only Taro came''); hereafter, \textit{dake} sentence) are less similar on the surface but semantically equivalent to \textit{shika} sentences.
To investigate whether LLMs capture semantic rather than superficial similarity, we compare
\begin{enumerate*}[label=(\roman*)]
    \item the similarity~($\simsn$) between a \textit{shika} sentence and its corresponding negative sentence and
    \item the similarity~($\simsd$) between a \textit{shika} sentence and its corresponding \textit{dake} sentence.
\end{enumerate*}
We extract \textit{shika} sentences from the 819 instances in \datasetname, removing duplicate entries.
Furthermore, to retain only instances with clear entailment relations, we filter out those containing entailment-canceling operators, resulting in 328 triples.
If the models capture semantic similarity rather than rely on superficial similarity, $\simsd$ is expected to be higher than $\simsn$.

\paragraph{Metrics}
To characterize the overall similarity, we compute the mean and standard deviation of $\simsd$ and $\simsn$.
Then, in order to assess whether $\simsn$ is significantly larger than $\simsd$.
We perform a Monte Carlo approximation to a paired permutation test with 10,000 random permutations.
We set the significance level to $\alpha = 0.001$.

\begin{table}[tbp]
    \centering
    \small
    \begin{tabular}{lcc}
    \toprule
    \multirow{2}{*}{Model} & \multirow{2}{*}{Mean (S.D.)} & Mean \\
     & &  Difference \\
    \midrule
    \multirow{2}{*}{Llama-3.1-8B} & $\simsd$: .954 (.010) & \multirow{2}{*}{0.014} \\
     & $\simsn$: .968 (.015) & \\
     \addlinespace
    \multirow{2}{*}{Llama-3.1-70B} & $\simsd$: .952 (.017) & \multirow{2}{*}{0.011} \\
     & $\simsn$: .963 (.033) & \\
     \addlinespace
    \multirow{2}{*}{Swallow-3.1-8B} & $\simsd$: .906 (.033) & \multirow{2}{*}{0.028} \\
     & $\simsn$: .933 (.040) & \\
     \addlinespace
    \multirow{2}{*}{Swallow-3.1-70B} & $\simsd$: .946 (.033) & \multirow{2}{*}{0.011} \\
     & $\simsn$: .957 (.033) & \\
     \addlinespace
    \multirow{2}{*}{LLM-jp-4-8B-It} & $\simsd$: .977 (.013) & \multirow{2}{*}{0.003} \\
     & $\simsn$: .980 (.015) & \\
    \bottomrule
    \end{tabular}
    \caption{Mean and standard deviation of $\simsd$ and $\simsn$ for each model, together with the mean difference used in the paired permutation test ($p<0.001$ for all models).}
    \label{tab:similarity_results}
\end{table}

\subsection{Results and Discussions}
\label{subsubsec:similarity_results}
The results are shown in \Cref{tab:similarity_results}.
The mean values of both $\simsd$ and $\simsn$ were at least 0.9 for all models, likely because there is an overlap of words in each sentence pair.
Comparing the two similarity measures, $\simsn$ consistently yields a higher mean than $\simsd$ across all models.
Moreover, paired permutation tests showed that the mean difference was significantly greater than zero for every model.
These results suggest that the embeddings of \textit{shika} sentences are consistently closer to their corresponding negative sentences than to their corresponding \textit{dake} sentences,
despite the latter being more semantically similar. 

This suggests that even highly accurate models rely more on superficial similarity than semantic relatedness in their internal representations.

\section{Conclusion}
\label{sec:conclusion}
We constructed \datasetname, an NLI dataset for studying LLMs' understanding of Japanese focus particles, through the semi-automated augmentation of a subset of JSeM with human validation.
Using \datasetname, we evaluated LLMs from the three perspectives: task performance, robustness, and confidence.

Our evaluation results show that
\begin{enumerate*}[label=(\roman*)]
    \item all models generally handled straightforward instances correctly;
    \item all models were robust to lexical variation;
    \item confidence scores were generally higher for correct than for incorrect predictions.
\end{enumerate*}

Nevertheless, all the models struggled with challenges, such as presupposition and superficially similar but semantically distinct premise--hypothesis pairs.
They also often produced incorrect predictions with high confidence.
Furthermore, analyses of model reasoning contents and internal representations suggest that these incorrect predictions might stem from an insufficient understanding of presupposition and an over-reliance on surface-level similarity.

We believe that \datasetname provides a useful testbed for evaluating and analyzing LLMs' understanding of Japanese focus particles.
In particular, it enables systematic investigation not only of their ability to capture the inference patterns associated with focus particles but also of their robustness to lexical variation.

\section*{Limitations}
\label{sec:limitations}
Our study has several limitations.
First, our experiments were conducted on a limited set of commercial, open-source, and Japanese-focused LLMs.
As there may be better-performing models and new models continue to emerge, the observations reported in this paper may not fully generalize to such LLMs.

Another limitation concerns the analysis of internal representations in \Cref{subsec:surface_similarity}.
We investigated only sentence pairs involving \textit{shika} when analyzing the similarity between superficially similar expressions.
For \textit{shika} sentences, it is possible to construct triplets in which the corresponding \textit{dake} sentence is semantically more similar to the original sentence, while the corresponding negative sentence is superficially more similar.
Such triplets provide a natural basis for comparing semantic and superficial similarity.
However, constructing comparable triplets for other focus particles is considerably more difficult.
Therefore, it remains unclear whether the observed tendency generalizes to other focus particles or related linguistic phenomena.

In addition, while our study treated token probability as a measure of the models' confidence in their responses, some previous studies suggest that this proxy might be inadequate.
In particular, \citet{tian-etal-2023-just} demonstrated that for LMs fine-tuned with reinforcement learning from human feedback, verbalized confidence scores can be better calibrated than confidence estimates derived from the models' conditional token probabilities.
Furthermore, \citet{xiong-etal-2024-can} showed that it remains difficult to derive reliable confidence estimates from LLMs using a single method.
Therefore, it may be possible to maximize reliability by combining multiple methods.

\section*{Ethical Considerations}
\label{sec:ethical_statement}
\datasetname is constructed by using JSeM, which is distributed under the BSD 3-Clause License.
We complied with the original license terms in constructing and distributing our dataset.
The additional examples were generated through manual and LLM-based augmentation and subsequently verified by human annotators.
The dataset contains only synthetic examples and does not include personal or sensitive information.

\section*{Acknowledgement}
We would like to thank the three anonymous reviewers and the meta-reviewer for their valuable comments and suggestions.
We also thank the human annotators for their contributions to improving the quality of \datasetname.
Finally, we are grateful to Zhihan Cao, Ryoma Kumon, Koki Ryu, and Taisei Yamamoto for their helpful feedback on the manuscript.
This work was supported by JST CREST Grant Number JPMJCR2565 and JST BOOST Program Grant Number JPMJBY24H5, Japan.

\bibliography{ref,anthology-1,anthology-2}

\appendix

\section{Gloss Abbreviations}
\label{apd:gloss_abbreviations}
The glossing abbreviations used in this paper follow the Leipzig Glossing Rules.
The abbreviations are listed below.
\begin{itemize}
    \item ACC: accusative
    \item DAT: dative
    \item GEN: genitive
    \item NEG: negation
    \item NOM: nominative
    \item NPST: non-past
    \item POSB: possibility
    \item PROG: progressive
    \item PRS: present
    \item PST: past
    \item TOP: topic
\end{itemize}

\section{Examples of Added Problems}
\label{apd:added_problems}
In this section, we introduce the additional instances added beyond those described in \Cref{subsec:new_inference_type}.
\paragraph{Presupposition by \textit{Dake-wa}}
Although \textit{dake} (``only'') alone is typically considered not to trigger a presupposition~\citep{ido2021hidden,oshima2023semantic}, it has been argued that \textit{dake} combined with \textit{wa} (``TOP'') functions as a presupposition trigger~\citep{oshima2023semantic}.
Therefore, we add an instance using (\ref{ex:dake-wa}) as the premise and its presupposition, ``No one other than Taro came,'' as the hypothesis.
\begin{exe}
    \ex\label{ex:dake-wa}
        \gll Taro-\textbf{dake-wa} ki-ta. \\
        Taro-only-\Top{} come-\Pst{} \\
        \glt ``Only Taro came.''
\end{exe}

\paragraph{\textit{Mo} and \textit{Demo} Attached to Interrogative Words}
When attached to interrogative words, \textit{mo} and \textit{demo} can give rise to a special interpretation.
For example, \textit{dare-mo} (who-mo) combined with negation conveys a universal negative meaning.
Therefore, when (\ref{ex:dare-mo}) is used as the premise and ``Someone is in the room'' is used as the hypothesis, the label is \textit{no}.
\begin{exe}
    \ex\label{ex:dare-mo}
        \gll Heya-ni-wa dare-\textbf{mo} i-nai. \\
        room-\Dat{}-\Top{} who-mo exist-\Neg{} \\
        \glt ``There is no one in the room.''
\end{exe}
In addition, \textit{nan-demo} (what-demo) can give rise to a universal interpretation.
Therefore, we construct an instance using (\ref{ex:nan-demo}) as the premise and ``Taro does not eat apples'' as the hypothesis, which receives the label \textit{no}.
\begin{exe}
    \ex\label{ex:nan-demo}
        \gll Taro-wa nan-\textbf{demo} tabe-ru. \\
        Taro-\Top{} what-demo eat-\Prs{} \\
        \glt ``Taro eats anything.''
\end{exe}
Furthermore, a sequence of an interrogative word, a classifier\footnote{A classifier is a word that follows a numeral to indicate the type of thing being counted.
For example, \textit{satsu} is often used to count the number of books, and \textit{3-satsu-no hon} (3-CLS-GEN book) means three books.}, and \textit{mo} indicates that several entities satisfy the predicate.
Therefore, we add an instance using (\ref{ex:nan-nin-mo}) as the premise and ``Only one person came'' as the hypothesis, which receives the label \textit{no}.
\begin{exe}
    \ex\label{ex:nan-nin-mo}
        \gll Hon-o nan-satsu-\textbf{mo} kat-ta. \\
        book-\Acc{} what-\Cls{}-mo buy-\Pst{} \\
        \glt ``I bought several books.''
\end{exe}

\paragraph{Minimal Quantifier Expressions with \textit{Mo} under Negation}
When \textit{mo} is attached to a minimal quantity expression, such as \textit{hitori} (``one person''), and occurs with negation, the resulting expression conveys universal negation.
Therefore, we add an instance using (\ref{ex:hitori-mo}) as the premise and ``Someone came'' as the hypothesis, which receives the label \textit{no}.
\begin{exe}
    \ex\label{ex:hitori-mo}
        \gll Hitori-\textbf{mo} ko-nakatta. \\
        one.person-mo come-\Pst{}.\Neg{} \\
        \glt ``No one came.''
\end{exe}

\section{Annotation Procedure and Compensation}
\label{apd:human_annotator}
We recruited three native Japanese speakers with no formal background in linguistics or semantics as annotators through the Japanese crowdsourcing platform Lancers.\footnote{https://www.lancers.jp/}
Before the annotation, we provided an overview of the task, together with a small number of example instances, to ensure that the annotators understood the annotation procedure.
Because our objective was to evaluate human judgments, we explicitly instructed annotators not to use AI-based tools during the annotation process.

The validation set consisted of 2,082 instances.
We estimated that each instance would require approximately 30 seconds to annotate, resulting in an estimated total workload of about 17 hours.
Compensation was calculated at a rate of 1,250 JPY per hour, which exceeded the minimum wage in Tokyo, Japan, at the time of annotation (April 2026).
Furthermore, the annotators were compensated for the full estimated workload of 17 hours, regardless of the actual time required to complete the task.

\section{Label Definitions for Validation}
\label{apd:label_definition}
We presented the definitions of the labels as follows:
\begin{itemize}
    \item \textit{entailment}: the case where the hypothesis is necessarily true given the premise
    \item \textit{contradiction}: the case where the hypothesis is necessarily false given the premise
    \item \textit{neutral}: the case that is neither entailment nor contradiction
    \item \textit{invalid}: the case where the premise or the hypothesis is ungrammatical or does not make sense
\end{itemize}
\textit{Entailment}, \textit{contradiction}, and \textit{neutral} correspond to \textit{yes}, \textit{no}, and \textit{unknown}, respectively.

\section{List of Focus Particles}
\label{apd:fp_list}
\begin{table}[htbp]
    \centering
    \begin{minipage}[t]{0.23\textwidth}
        \centering
        \begin{tabular}{ll}
        \toprule
            Particle & Count \\
        \midrule
            shika & 819 \\
            dake & 807 \\
            nomi & 546 \\
            bakari & 284 \\
            kagiri & 35 \\
            mo & 584 \\
            sae & 225 \\
            sura & 292 \\
            made & 292 \\
        \bottomrule
        \end{tabular}
    \end{minipage}
    \begin{minipage}[t]{0.23\textwidth}
        \centering
        \begin{tabular}{ll}
        \toprule
            Particle & Count \\
        \midrule
            wa & 340 \\
            toka & 466 \\
            nado & 627 \\
            nanka & 125 \\
            nante & 243 \\
            koso & 80 \\
            kurai & 155 \\
            hodo & 202 \\
            demo & 99 \\
        \bottomrule
        \end{tabular}
    \end{minipage}
    \caption{Focus particles in the expanded dataset.}
    \label{tab:particle_list}
\end{table}

\Cref{tab:particle_list} shows the list of Japanese focus particles on which we focus in this study.
The \textit{Count} column indicates the number of instances in \datasetname involving each focus particle.

\textit{Shika}, \textit{dake}, \textit{nomi}, \textit{bakari}, and \textit{kagiri} generally express exclusivity.
\textit{Mo}, \textit{sae}, \textit{sura}, and \textit{made} primarily express additivity.
\textit{Wa} expresses contrast, whereas \textit{koso} expresses emphasis.
\textit{Kurai} and \textit{hodo} are used to indicate degree.
\textit{Demo}, \textit{toka}, \textit{nado}, and \textit{nanka} have multiple usages but are primarily used for exemplification.
Likewise, \textit{nante} has multiple usages and is often used to express surprise, among other functions.

\section{Model URLs}
\label{apd:model_url}
The URLs of the models used in the experiments are listed below.
\begin{itemize}
    \item Swallow-3.1 (v0.1): \url{https://huggingface.co/collections/tokyotech-llm/llama-31-swallow}
    \item LLM-jp-4-8B: \url{https://huggingface.co/collections/llm-jp/llm-jp-4-models}. For the thinking model, we set the reasoning effort to \textit{medium}.
    \item Llama-3.1: \url{https://huggingface.co/collections/meta-llama/llama-31}
    \item GPT-5.2 (used in the default reasoning effort): \url{https://openai.com/ja-JP/index/gpt-5-system-card}
\end{itemize}

\section{Experimental Prompt}
\label{apd:prompt}
\begin{table*}[htbp]
    \centering
    \begin{tabular}{ll}
    \toprule
        System &
        \begin{minipage}[t]{0.8\linewidth}
            以下に示す前提に対する仮説の関係が含意ならyes，矛盾ならno，中立ならunknownと答えてください．yes, no, unknown以外の解答や説明はしないでください．\\
            (``Given the premise and hypothesis below, determine whether the hypothesis is entailed by the premise. Answer yes for entailment, no for contradiction, and unknown for neutral. Output only yes, no, or unknown, and do not provide any explanation.'')
        \end{minipage}\\
    \midrule
        User &
        \begin{minipage}[t]{0.8\linewidth}
            前提1：<premise 1>\\
            前提2：<premise 2>\\
            ...\\
            仮説：<hypothesis>\\
            解答：\\
            (``Premise 1: <premise 1> \\
            Premise 2: <premise 2>\\
            ...\\
            Hypothesis: <hypothesis>\\
            Answer:'')
        \end{minipage}\\
    \bottomrule
    \end{tabular}
    \caption{System and user prompts for non-thinking models. When using the standard NLI labels, we replaced \textit{yes}, \textit{no}, and \textit{unknown} with \textit{entailment}, \textit{contradiction}, and \textit{neutral}, respectively.}
    \label{tab:prompts}
\end{table*}

\Cref{tab:prompts} shows the prompt used in our experiments for non-thinking models.
For LLM-jp-4-8B-Think, we used the same prompt except that we removed the instruction, ``do not provide any explanation.''

\section{Accuracy on Instances Regarding Presupposition Associated with Each Particle}
\label{apd:presupposition_per_particle}
\begin{table*}[tbp]
    \centering
    \begin{tabular}{lcccccccccc}
    \toprule
    Model & shika & dake & mo & sae & sura & made & toka & nado & koso & kurai\\
    \midrule
    Llama-3.1-8B & .687 & .670 & .470 & \textbf{.423} & .472 & .321 & .685 & .750 & .500 & .500 \\
    Llama-3.1-70B & .476 & .670 & \textbf{.530} & .319 & .422 & .294 & .202 & .259 & \textbf{1.000} & .600 \\
    Swallow-3.1-8B & \textbf{.837} & .990 & .518 & .399 & \textbf{.550} & .440 & \textbf{.807} & .780 & \textbf{1.000} & .600 \\
    Swallow-3.1-70B & .482 & .734 & .185 & .092 & .132 & .074 & .442 & .436 & \textbf{1.000} & .700 \\
    LLM-jp-4-8B-It & .669 & .979 & .134 & .077 & .238 & .220 & .584 & .677 & .000 & .700 \\
    LLM-jp-4-8B-Think & .470 & \textbf{1.000} & .054 & .033 & .007 & .243 & .752 & \textbf{.811} & .000 & .300 \\
    GPT-5.2 & .542 & .830 & .470 & .283 & .403 & \textbf{.449} & .337 & .375 & \textbf{1.000} & \textbf{.800} \\
    \midrule
    Count & 83 & 47 & 168 & 168 & 160 & 148 & 119 & 116 & 8 & 5\\
    \bottomrule
    \end{tabular}
    \caption{Accuracy and APA of each model on problems involving presupposition for each particle in the zero-shot setting. Each value represents the average of the results obtained when each model was prompted to provide either JSeM-compliant labels (\textit{yes}, \textit{no}, \textit{unknown}) and when it was prompted to provide standard NLI labels (\textit{entailment}, \textit{contradiction}, \textit{neutral}). }
    \label{tab:presup_per_particle}
\end{table*}

\Cref{tab:presup_per_particle} shows each model's accuracy on instances regarding presupposition associated with each particle.
All models exhibit a substantial drop in accuracy on the particles \textit{mo}, \textit{sae}, \textit{sura} and \textit{made} compared to their overall accuracy.
A common property of these particles is that they all trigger an additive presupposition (e.g., ``Someone other than Taro came'' for \textit{Taro-mo ki-ta} (``Taro also came'')).
Given that the additive presupposition triggered by \textit{mo} is relatively robust, the low accuracy suggests that these models have difficulty adequately capturing presupposition.

\section{Analysis of Thinking Model's Outputs}
\label{apd:thinking_outputs}
We examined the reasoning outputs of LLM-jp-4-8B-Think in the zero-shot setting and found that the model tended to judge presuppositional content as not being entailed.
For example, given a premise such as \textit{Taro-sae repoto-o dasi-ta} (``Even Taro submitted a report'') and a hypothesis (Someone other than Taro submitted a report), the model produced the following analysis: ``The premise implies that Taro submitted a report; it does not specify whether others did or not.''
However, the focus particle \textit{sae} (''even'') presupposes the existence of an alternative individual for whom the proposition also holds.
This reasoning output suggests that the model fails to capture the presupposition triggered by the focus particle \textit{sae}.

Furthermore, even when the model correctly identified the presupposition, it sometimes treated it as being canceled by modal expressions, which should not happen.
For example, the sentence \textit{Kyaku-wa Ichiro-shika ko-nakatta kamoshirenai} (``Perhaps only Ichiro came as a customer'') presupposes that Ichiro came as a customer, and this presupposition is not canceled by the modal expression.
However, the model produced the following analysis: ``Premise says maybe only Ichiro came as a customer; it's possible but not certain. The premise does not guarantee that Ichiro is a customer; it just says perhaps only Ichiro came.''
This indicates that the model incorrectly treated the presupposition as falling within the scope of the modal operator.

These suggest that the model's errors stem from being overly conservative in drawing inferences from presuppositional content and from incorrectly treating presuppositions as being canceled by entailment-canceling operators.

\section{Models' Performance on Downstream QA Tasks}
\label{apd:downstream}
We investigate whether the inference patterns that the models struggle with under the zero-shot setting also affect their performance on downstream question-answering (QA) tasks.
Specifically, we constructed QA examples from 45 \datasetname instances involving the presupposition triggered by \textit{mo}.
For example, given the premise \textit{Taro-mo repoto-o dasi-ta} (``Taro also submitted a report''), we ask, \textit{Taro-igai-ni repoto-o dasi-ta hito-ga imasu-ka?} (``Is there anyone other than Taro who submitted a report?'').
In this case, the gold label is \textit{yes}.

The results are shown in \Cref{tab:qa_results}.
The models other than Llama exhibit particularly low accuracy.
These results suggest that the inability to correctly capture presuppositions in NLI tasks may also affect performance on downstream tasks.
\begin{table}[tbp]
    \centering
    \begin{tabular}{lc}
    \toprule
    Model & Accuracy\\
    \midrule
    Llama-3.1-8B & .644 \\
    Llama-3.1-70B & .844 \\
    Swallow-3.1-8B & .444 \\
    Swallow-3.1-70B & .378 \\
    LLM-jp-4-8B-It & .467 \\
    LLM-jp-4-8B-Think & .089 \\
    GPT-5.2 & .089 \\
    \bottomrule
    \end{tabular}
    \caption{Accuracy of each model on QA instances involving presupposition of \textit{mo}.}
    \label{tab:qa_results}
\end{table}

\section{Confidence and Earliest Decisive Layer}
\label{apd:conf_and_edl}
\Cref{subsec:logit_confidence} examined the relationship between model confidence and prediction correctness.
As a supplementary analysis, we investigate the relationship between model confidence and the stage at which a model's prediction becomes stable.
Intuitively, humans tend to arrive at confident decisions early.
We therefore investigate whether LLMs exhibit a similar tendency.

\subsection{Settings}
\label{subsubsec:conf_and_edl_settings}
We analyze the logit trajectory to obtain the provisional prediction at each layer.
Specifically, at each layer, we compare the logits of the candidate labels and regard the label with the highest logit as the layer-wise prediction.
We define the earliest decisive layer as the earliest layer at which the layer-wise prediction matches the final prediction and remains unchanged for all subsequent layers.
We measure the relationship between prediction confidence and the earliest decisive layer using Spearman's rank correlation coefficient.

\subsection{Results and Discussions}
\label{subsubsec:conf_and_edl_results}
\begin{table}[tbp]
    \centering
    \begin{tabular}{lc}
    \toprule
    Model & Spearman's $\rho$ \\
    \midrule
    Llama-3.1-8B & -.425 \\
    Llama-3.1-70B & -.772 \\
    Swallow-3.1-8B & -.175 \\
    Swallow-3.1-70B & -.586 \\
    LLM-jp-4-8B-It & -.676 \\
    \bottomrule
    \end{tabular}
    \caption{Spearman correlation coefficients between prediction confidence and the earliest decisive layer for each model. All correlations are significant at $p < 0.001$.}
    \label{tab:conf_and_edl_results}
\end{table}

The results are shown in \Cref{tab:conf_and_edl_results}.
All models except Swallow-8B exhibit moderate to strong negative correlations, while Swallow-8B also shows a weak negative correlation.
These results suggest that for most models, predictions made with higher confidence tend to become fixed at earlier layers.
Viewing the logit trajectory as a temporal process, this finding suggests that the models can commit to their predictions earlier when they are more confident.

\section{Use of AI Assistants}
We used ChatGPT\footnote{\url{https://chatgpt.com}} and Gemini\footnote{\url{https://gemini.google.com}} for writing refinement, including improving the fluency of English text and discussing alternative phrasings.
These tools were not used to generate the scientific content, experimental design, analyses, or conclusions of this paper.
All technical content, experimental procedures, results, and interpretations were developed and verified by the authors, who take full responsibility for the final manuscript.

\end{document}